\documentclass[conference]{IEEEtran}
\IEEEoverridecommandlockouts
\usepackage{cite}
\usepackage{multirow}
\usepackage{amsmath,amssymb,amsfonts}
\usepackage{algorithmic}
\usepackage{graphicx}
\usepackage{textcomp}
\usepackage{xcolor}
\def\BibTeX{{\rm B\kern-.05em{\sc i\kern-.025em b}\kern-.08em
    T\kern-.1667em\lower.7ex\hbox{E}\kern-.125emX}}
\begin{document}

\makeatletter
\newcommand{\linebreakand}{%
  \end{@IEEEauthorhalign}
  \hfill\mbox{}\par
  \mbox{}\hfill\begin{@IEEEauthorhalign}
}
\makeatother

\bibliographystyle{plain}

\title{Adversarial Learning of Hard Positives for Place Recognition 
}

\author{\IEEEauthorblockN{1\textsuperscript{st}Wenxuan Fang \thanks{\IEEEauthorrefmark{1} These authors contributed equally to this work.}\IEEEauthorrefmark{1}}
\IEEEauthorblockA{\textit{Shenzhen International Graduate School} \\
\textit{Tsinghua University}\\
China \\
fwx20@mails.tsinghua.edu.cn}

\\
\IEEEauthorblockN{3\textsuperscript{rd} Yoli Shavit}
\IEEEauthorblockA{\textit{Faculty of Engineering} \\
\textit{Bar Ilan University}\\
Israel \\
yolisha@gmail.com}
\and
\IEEEauthorblockN{2\textsuperscript{nd} Kai Zhang \IEEEauthorrefmark{1}}
\IEEEauthorblockA{\textit{Shenzhen International Graduate School} \\
\textit{Tsinghua University}\\
China \\
zhangkai@sz.tsinghua.edu.cn}
\\
\IEEEauthorblockN{4\textsuperscript{th} Wensen Feng \thanks{\IEEEauthorrefmark{2} Corresponding author.}\IEEEauthorrefmark{2}}
\IEEEauthorblockA{\textit{Central Media Technology Institute} \\
\textit{HuaWei Technologies}\\
China \\
fengwensen@huawei.com
}
}

\maketitle

\begin{abstract}
Image retrieval methods for place recognition learn global image descriptors that are used for fetching geo-tagged images at inference time. Recent works have suggested employing weak and self-supervision for mining hard positives and hard negatives in order to improve localization accuracy and robustness to visibility changes (e.g. in illumination or view point). However, generating hard positives, which is essential for obtaining robustness, is still limited to hard-coded or global augmentations. In this work we propose an adversarial method to guide the creation of hard positives for training image retrieval networks. Our method learns local and global augmentation policies which will increase the training loss, while the image retrieval network is forced to learn more powerful features for discriminating increasingly difficult examples. This approach allows the image retrieval network to generalize beyond the hard examples presented in the data and learn features that are robust to a wide range of variations. Our method achieves state-of-the-art recalls on the Pitts250 and Tokyo 24/7 benchmarks and outperforms recent image retrieval methods on the rOxford and rParis datasets by a noticeable margin.

\end{abstract}


\section{Introduction}
\begin{figure}[h]
\centering 
\includegraphics[width=0.5\textwidth]{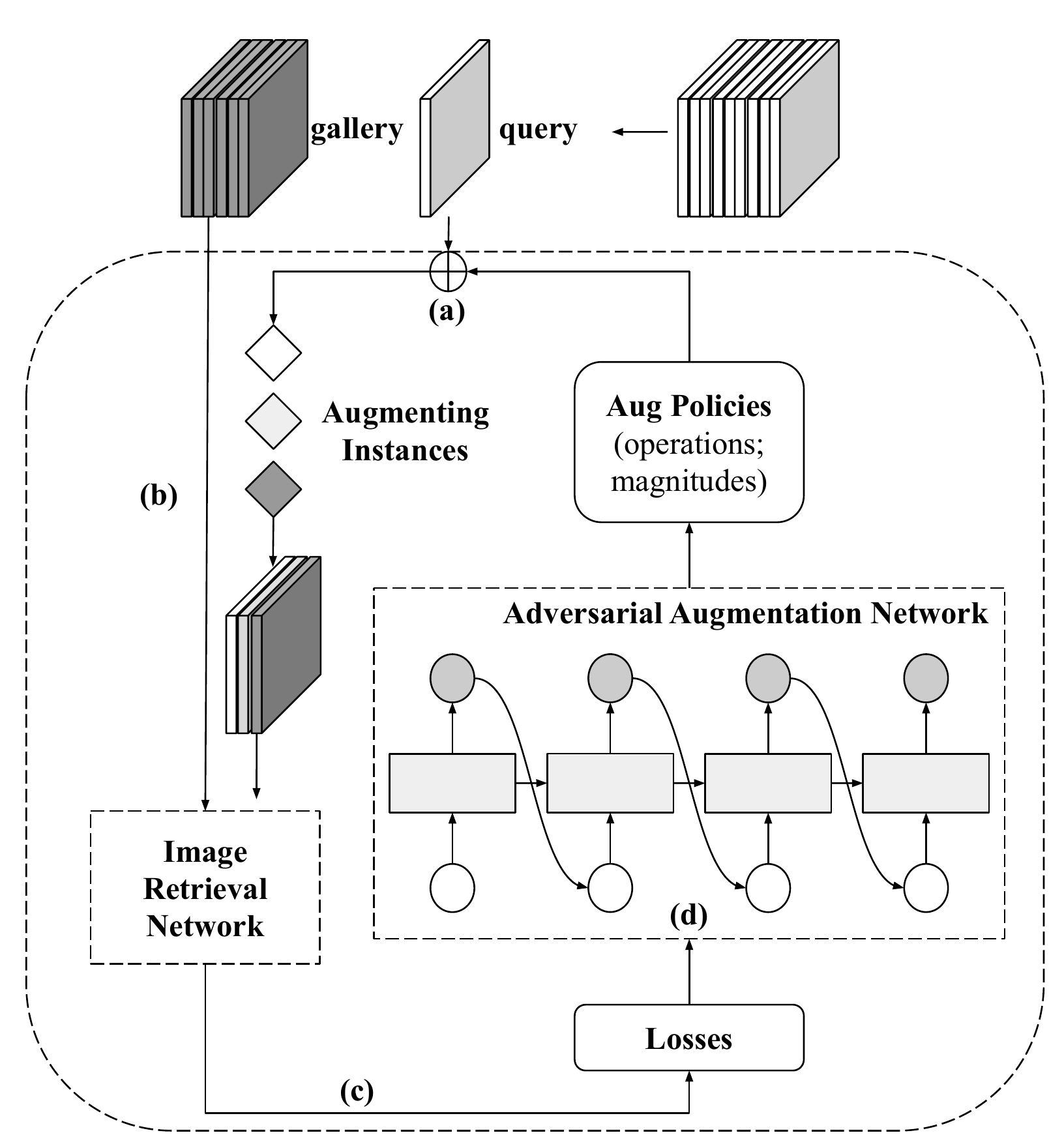} 
\caption{An illustration of our proposed method. Query images are augmented according to learned augmentation policies (a), while reference images are processed with standard operations (b). An IR network is then trained to minimize a contrastive loss over query and database images (c). Constrastive losses corresponding to different policies guide an adversarial augmentation network to maximize them by generating new adversarial policies.} 
\label{Fig.1} 
\end{figure}

One key challenge in place recognition is to learn image representations with sufficient discriminative power for distinguishing between repetitive and similar locations in a GPS-labeled dataset \cite{aug8}. In recent years, Convolutional Neural Networks (CNNs), together with different pooling operations, have been extensively used for this purpose \cite{netvlad, finetunegem,noh2017large,gu2018attention,teichmann2019detect,husain2019remap,delg}These networks are typically trained with contrastive learning \cite{netvlad, aploss, sare, sfrs}, encouraging images from the same location to be summarized into similar representations and vice versa. These methods rely on mining strategies for positive and negative examples, where the network is typically gradually faced with harder examples, obtained based on GPS tagging \cite{BoW,finetunegem,sare,sfrs}. However, generating hard positives is usually limited to a small set of images that are visually similar to the query image but are captured at a different location. Defining this set is not trivial and either relies on the network itself or on some manual pre-processing.
Additionally generating hard positives  from the image itself is typically implemented with hard-coded augmentations which require expert knowledge \cite{handf1, handf2, handf3}.

In order to address these limitations we propose an adversarial method for training image retrieval (IR) networks (Fig. \ref{Fig.1}). Inspired by learned data augmentation approaches \cite{aug2, aug3, aug4, aug5, aug1, aug6}, our adversarial agent (named 'adversarial augmentation network', Fig. \ref{Fig.1}-a,d) searches for augmentation policies which will produce gradually harder positive examples.The target IR network, acting as a discriminator, will aim to correctly classify these challenging examples Fig. \ref{Fig.1}-c). Consequently, instead of training an IR network with a fixed dataset \cite{BoW,finetunegem,sare,sfrs} our method will directly generate difficult positive examples that\textit{do not appear} in the original data and that are dynamically adjusted to enforce learning of more robust and powerful features.  

 We validate the effectiveness of our method by evaluating it on several large-scale place recognition datasets. Our method achieves state-of-the-art recalls on both the Tokyo 24/7 \cite{tokyo} and the Pitts250 \cite{pitts} benchmarks, with a 0.9\%-2.1\% improvement compared to the current leading IR method (SFRS \cite{sfrs}). In addition, our method also generalizes well on the rOxford and rParis datasets, surpassing current state-of-the-art methods by a noticeable margin.
 
In summary, our contributions are as follows:
\begin{itemize}
    \item We propose an adversarial framework for training IR networks. In order to counter harder examples generated by adversarial augmentation strategies, the target IR network must learn robust features with improved generalization power.
     \item The proposed approach  can directly generate difficult positive samples that do not appear in the original dataset and dynamically adjust them based on the optimization process of the IR network. 
    \item Our method achieves state-of-the-art recalls on challenging large-scale place recognition benchmarks (Tokyo24/7,  Pitts250) and demonstrates improved generalization on the  rOxford and rParis datasets. 
\end{itemize}

\section{Related Work}
\paragraph{Mining Hard Examples for Training IR Networks} 
In the past decade, CNNs were shown to successfully learn visual features and their aggregation, replacing their hand-crafted counterparts such as a Bag of Words (BoW)\cite{BoW}, Fisher Vectors (FV)\cite{fisher}, or Vector of Locally Aggregated Descriptors (VLAD) \cite{vlad1} and becoming the standard backbone for IR networks. 
In recent years, the focus has gradually shifted from designing CNN based architectures and pooling operations \cite{netvlad}, \cite{finetunegem},\cite{delg} to employing weak/self supervision and contrastive learning for training IR networks. Specifically, recent works have focused on training and optimization strategies for IR networks, including the use of triplet loss, contrastive loss and average precision loss \cite{netvlad,aploss}, as well as different mining strategies required for realizing these losses \cite{sare, sfrs}. Self-supervising Fine-grained Region Similarities (SFRS) \cite{sfrs} further investigates multi-scale contextual information by refining image regions.
Our work follows a similar motivation of creating increasingly difficult positive samples for visual geo-localization in order to encourage the learning of robust image representations.
However, as opposed to SFRS\cite{sfrs}, which relies on GPS tagging and on global and local similarity mining, our method samples augmentation policies which operate at the image level and are dynamically adjusted to gradually challenge the IR network.

\paragraph{Data Augmentation} The main motivation for using data augmentation is to improve the generalization of the learned model and the robustness of learned representations to variations and noise. 

Traditional handcrafted augmentation techniques include simple transformations such as horizontal flips, color space enhancement, and random cropping. However, the augmentation policy required for optimizing the image representation is not known and is often manually set based on some assumptions. In the past few years, deep learning-based data augmentation methods have been proposed, employing feature space augmentation, adversarial training, and generative adversarial networks are being expanded.
AutoAugment \cite{autoaug} creates a search space for augmentation strategies, and selects appropriate sub-strategies (such as clipping, flipping, etc.) by designing the operation of the search algorithm. With a similar motivation in mind, Wang et al.\cite{daas} proposed to apply a differentiable network architecture search to augmentation strategy search. Instead of maintaining an augmentation search space, RandAugmentation\cite{random} suggested to assemble augmentation policies by randomly sampling augmentation operations, offering a lightweight alternative which can be directly integrated into the training process of the target model. In our work, we utilize a recurrent neural network (RNN) to sample data augmentation strategies and train it together with the target model (IR network) in an adversarial learning setup.

\section{Method}

The key to our approach is to increase data diversity by gradually generating more adversarial examples during the training process where an augmentation policy generator attempts to maximize the loss of the IR network, which in turn is optimized to minimize it. In the following sections we describe the three main components of our approach: (a) an IR network (b) an adversarial augmentation network and (c) the proposed adversarial learning framework.

\subsection{Image Retrieval Network}

For our target IR network we employ the architecture used by SFRS\cite{sfrs}. 
We describe this architecture below for completeness. The network consists of a VGG16 backbone \cite{vgg} and a VLAD layer \cite{netvlad} which encodes and aggregates feature maps into a compact representation. In addition to global image features, the VLAD layer is also applied on features derived from sub-image regions. Specifically, feature maps from the convolutional backbone are transformed into four quarter sub-regions (top-left, top-right, bottom-left, bottom-right quarters) before being fed into the VLAD layer. We then sum the residuals and obtain four-quarter vectors $ V_i^1,V_i^2,V_i^3,V_i^4 $ corresponding to the four sub-regions. As illustrated in Fig. \ref{Fig.2}, the four quarter vectors are then combined to form half region and global image features. The resulting matrix $V$ is then intra-normalized, aggregated into a vector and L2-normalized. 

\begin{figure}[h]
\centering 
\includegraphics[width=0.5\textwidth]{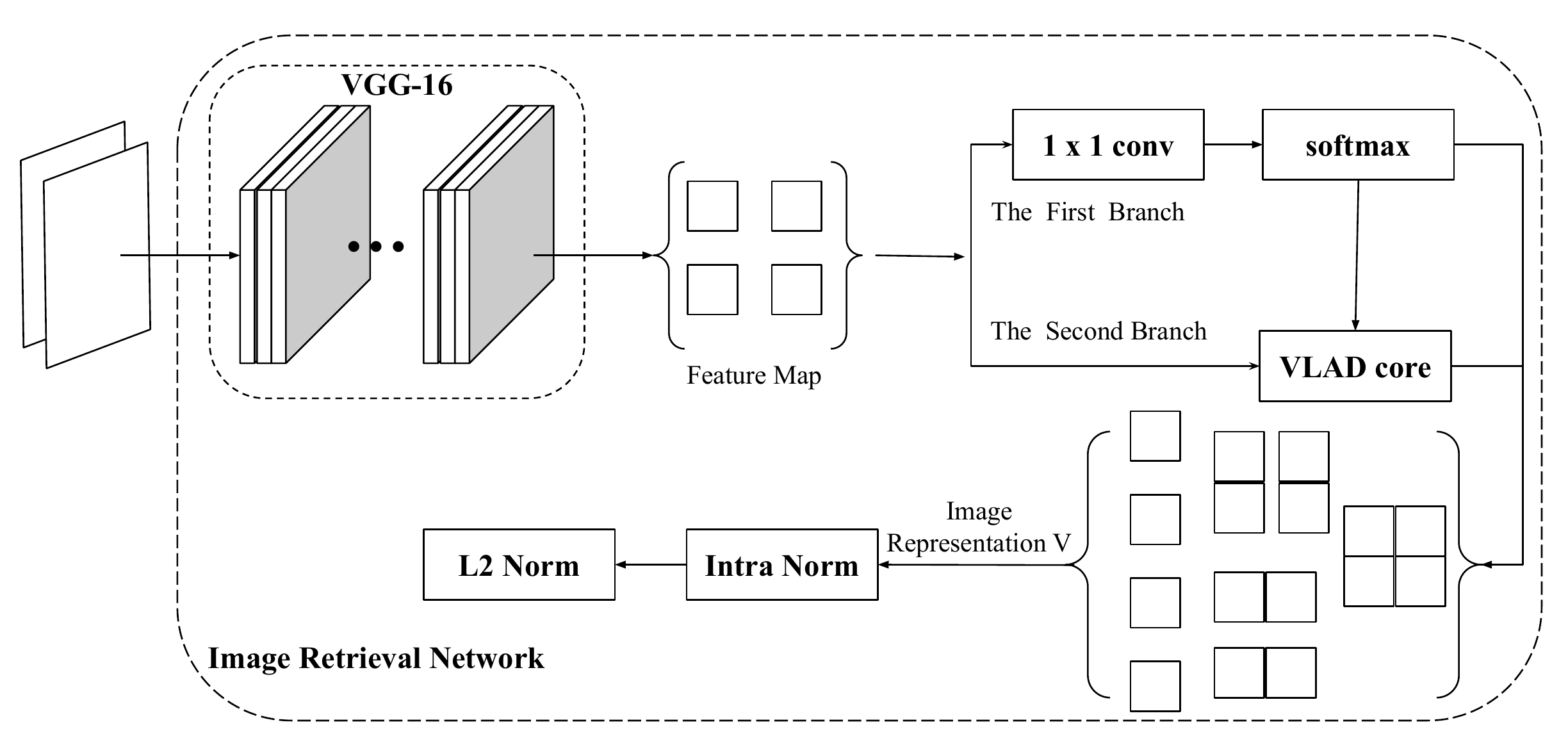} 
\caption{The architecture of our IR network.} 
\label{Fig.2} 
\end{figure}

\subsection{Adversarial Augmentation Network}
For generating augmentation policies we require a search space and a search algorithm. 
\subsubsection{Search space} We include 14 different operations in our search space including: ShearX/Y, TranslateX/Y, Rotate, AutoContrast, Invert, Equalize, Solarize, Posterize, Color, Brightness, Sharpness and Sample Pairing. Each operation is assigned with a default range of magnitudes which are evenly discretized into two values. In order to ensure convergence in the adversarial learning process, the amplitude of all operations is set in a moderate range. In addition, randomness in the training process is introduced into our search space. Therefore, the policy search space in each epoch has $ \left(14\times10\right)^{10} $ possibilities. We define an augmentation policy as a collection of five sub-strategies, each containing two operations (to be sequentially applied). Each operation is assigned with an amplitude and a probability for applying it.

\subsubsection{Search algorithm} 
We use Reinforcement Learning as the search algorithm, which is implemented by a controller and a training method. The controller is implemented with a RNN architecture. At each time step of the RNN controller, a softmax layer will predict the action corresponding to a sub-policy and then the embedded predicted action will be sent to the next time step. The controller produces a total of five strategies from which we randomly choose one policy. 
The chosen policy is applied to augment the inputs used for training the IR network. The loss from the IR network is used to update the controller. This update is performed similarly to AutoAugment \cite{autoaug}, using the Policy Optimization algorithm.

\subsection{Adversarial Learning}
In this section we first describe the supervision process for our IR and Adversarial Augmentation Networks, respectively. 
We then set these processes within an adversarial learning framework.

\subsubsection{Supervising the IR Network} In order to make full use of the potential of difficult negative samples, the query-database image similarities are measured through self-supervision. We use $\theta_w$ to indicate the network parameters in $w$ different generations. For each query image, its $k$-nearest neighbors $p_1,…,p_i$ are first obtained by calculating Euclidean distances from the database image set. The query-database image similarities can be measured by a softmax function.

\begin{equation}
\begin{split}
S_{\theta_w}(q,p_1,\cdots,p_i;\tau _w)=softmax([\frac{<V_{\theta_w}^{q},V_{\theta_w}^{p_1}>}{ \tau_w },\\
\frac{<V_{\theta_w}^{q},V_{\theta_w}^{1_1}>}{ \tau_w },\cdots,
\frac{<V_{\theta_w}^{q},V_{\theta_w}^{1_4}>}{ \tau_w },\cdots,\\ \frac{<V_{\theta_w}^{q},V_{\theta_w}^{p_i}>}{ \tau_w } \frac{<V_{\theta_w}^{q},V_{\theta_w}^{i_1}>}{ \tau_w },\cdots, \frac{<V_{\theta_w}^{q},V_{\theta_w}^{i_4}>}{ \tau_w }])
\end{split}
\end{equation}
where $ V_{\theta_w}^{p_i} $ is the encoded feature representations of the $i$-th database image from the $w$-th generation network, $\tau_w$ is a hyper-parameter for adjusting the smoothness of similarity vector, and $ <·,·i> $ denotes dot product between two vectors. The relative similarity vectors $S_{\theta_w}$ estimated by the $(w-1)$-th generation network are used to supervise the $w$-th generation network via a “soft” cross-entropy loss:
\begin{equation}
\begin{split}
L_s(\theta_w) =\l(S_{\theta_w}(q,p_1,\cdots ,p_i;1),\\S_{\theta_{w-1}}(q,p_1,\cdots ,p_i;\tau_{w-1}))
\end{split}
\end{equation}
where $ \l(S,\hat{S})=-\sum_{i}^{}\hat{S}(i) log(S(i))\ $denotes the cross-entropy loss.

To mine the most difficult negative regions for each query image, we use a softmax-based hard loss, which computes the ratio between the query-positive regions against multiple query-negative regions:
\begin{equation}
L_h(\theta_w) =-\sum_{J=1}^{N}log\frac{e^{<V_{\theta_w}^{q},V_{\theta_w}^{p*}>}}{e^{<V_{\theta_w}^{q},V_{\theta_w}^{p*}>}+e^{<V_{\theta_w}^{q},V_{\theta_w}^{n_{j}^{*}}>}}.\
\end{equation}
In each iteration, both the “soft” cross-entropy loss and the original hard loss are jointly used for supervising the feature learning as follows:
\begin{equation}
L\left(\theta_w\right)=L_h(\theta_w)+\alpha L_s(\theta_w)\
\end{equation}
where $\alpha$ is the factor of the loss weight. The training process of every generation is performed by self-supervising query-database region similarities.

Considering the image retrieval network $I(·, \theta)$ with a loss function $L[I(x, \theta)]$, where each query image is transformed by data augmentation $\rho$ with the help of the augmentation policy network, the learning process of the image retrieval network can be defined as the following minimization problem.
\begin{equation}
\label{minloss}
arg \underset{\theta}{min}\underset{x\sim T}{E}\underset{\rho \sim A(\cdot,w)}{E}L[I(\rho(x_q),\theta)
\end{equation}

where T is the training set, x is the input image.$x_q$ is the query image of input samples. And $\rho$ represents the augmentation policy generated by network $A(\cdot,w)$.

\subsubsection{Supervising the Adversarial Augmentation Network}
 
Inspired by AutoAugment\cite{autoaug}, the augmentation policy network is designed to increase the training loss of the target network with harder augmentation policies. Therefore, we can mathematically express the object as the following maximization problem.
\begin{equation}
\label{maxloss}
arg \underset{\theta}{max}\underset{x\sim T}{E}\underset{\rho \sim A(\cdot ,w)}{E}L[I(\rho(x_q),\theta)]      
\end{equation}

\subsubsection{Adversarial Formulation}
In the context of adversarial learning, we can view the Adversarial Augmentation Network as a generator and IR Network as a discriminator, since they maximize and minimize the same loss (Eqs. \ref{maxloss} and \ref{minloss}, respectively), forming a closed loop of adversarial learning. Our model is trained for multiple generations, where each output of each generation is used for supervising the next generation. Specifically, after the training of the $i$-th generation converges, the $i+1$-th generation model is established and initialized. The $i$-th generation model is then fixed and performs similarity label estimation to supervise the training of the $i+1$ generation, forming a self-supervised process. The adversarial learning of Image Retrieval Network and Adversarial Augmentation Network is summarized as Algorithm 1.

\begin{table}
\centering
\begin{tabular}{lllll} 
\cline{1-1}
\textbf{Algorithm 1} Proposed Adversarial Learning Framework                                                                                                                                                                      &  &  &  &   \\ 
\cline{1-1}
\begin{tabular}[c]{@{}l@{}}\textbf{Input: }train epochs M, learning rate, generations G, input examples x, \\$x_q$ is the query image of input samples, $D$ different  instances of each \\ input augmented by polices, and $L_d$ refers to the loss generated by the \\ d-th policy\end{tabular} &  &  &  &   \\
\textbf{Initialization:} IR network $I(\cdot,\theta)$, augmentation network $A(\cdot,\omega)$\textbf{}                                                                                                                                                                &  &  &  &   \\
\textbf{for} g = 1 to G \textbf{do}                                                                                                                                                                                                                 &  &  &  &   \\
~ ~ ~\textbf{for} m = 1 to M \textbf{do}                                                                                                                                                                                                            &  &  &  &   \\
~ ~ ~ ~ ~$L_h=0,\forall h\in {\left \{ 1,2,\cdots,D \right \}}$                                                                                                                                                                    &  &  &  &   \\
~ ~ ~ ~ ~${\left \{ 1,2,\cdots,D \right \}}\leftarrow A(\cdot,\omega)$                                                                                                                                                       &  &  &  &   \\
~ ~ ~ ~ ~ \textbf{for} t=1 to T \textbf{do} ~ ~                                                                                                                                                                                                    &  &  &  &   \\
~ ~ ~ ~ ~ ~ ~ ~$\rho (x_q)\leftarrow (x_q)$                                                                                                                                                                                       &  &  &  &   \\
~ ~ ~ ~ ~ ~ ~ ~$L_d \leftarrow L(I(\rho (x_q),x,\theta))$                                                                                                                                                                       &  &  &  &   \\
~ ~ ~ ~ ~ ~ ~ ~$\theta \leftarrow \theta - \eta \frac{\partial L_d}{\partial \theta}$                                                                                                                                       &  &  &  &   \\
~ ~ ~ ~ ~~$L = \frac{1}{H}\sum_{h=1}^{H} L_d$                                                                                                                                                                                   &  &  &  &   \\
~ ~ ~ ~ ~~$\omega \leftarrow \omega - \eta \frac{\partial L}{\partial \omega}$                                                                                                                                            &  &  &  &   \\
Output~$\theta^{*},\omega^{*}$                                                                                                                                                                                            &  &  &  &   \\
\cline{1-1}
\end{tabular}
\end{table}

\section{Experimental Results}
\subsection{Implementation Details}
\paragraph{Datasets}
We validate our proposed method for place recognition using two challenging public image datasets: the Pittsburgh dataset \cite{pitts} and the Tokyo dataset \cite{tokyo}.
Pittsburgh (Pitts250k) is composed of around 24k query images and 250k database images which provides multiple street-level panoramic images taken at different times at close-by spatial locations from Google Street View\cite{2}. For a fair comparison with\cite{sfrs}, we followed the same division. The dataset is split into train, validation, and test sets, which was done geographically to ensure the sets contain independent images. In total, we end up with Pitts30k-train that contains 7,416 query images and 10,000 database images. For the validation, we use Pitts30k-val which consists of 7,608 query images and 10,000 database images. In the large-scale Pitts250k-test, we achieve the result of experiments which has 8,280 query images and 83,925 database images.

Tokyo 24/7 is another widely used benchmark in image localization. It provides 315 query images and 76k database images. Compared with Pittsburgh, it is more challenging. The challenge lies in vast illumination changes within the query images and between query and database images. Specifically, the query images were taken at daytime, sunset, and night, while database images were captured only during the daytime.

In addition, we also consider the
rOxford and rParis datasets, which are referred as the new annotations for the Oxford\cite{ro} and Paris\cite{rp} commonly benchmark datasets. These benchmarks define new and more difficult queries with challenging distractors. 
\paragraph{Evaluation}
We follow the standard place recognition evaluation meric of \cite{sfrs}. For the evaluating performance on the Pitts250k-test and 24/7 Tokyo, we use the percentage of Recall that can recognize query images correctly. For the image-retrieval datasets rOxford\cite{ro} and Paris\cite{rp}, the classical mean-Average-Precision (mAP) is used to measure landmark retrieval performance.
\paragraph{Training details}
We follow the same architecture and supervision details of SFRS for our IR Network. We use stochastic gradient descent (SGD) optimizer with momentum 0.9 and a constant learning rate of 0.001. 

Our adversarial augmentation network is implemented as a one-layer LSTM where the hidden unit size is set to 100, the embedding layer size is set to 32. We use the Proximal Policy Optimization algorithm to train the controller with Adam and an initial learning rate of 0.00035. In order to avoid unexpected fast convergence, we apply an entropy penalty with a weight of 0.00001. The selected augmented policy is applied in addition to the standard baseline pre-processing. For a query image, we first apply a baseline augmentation used by existing IR method. We then apply the operations and magnitudes of the selected policies.

Overall, the whole network is trained for four generations, where each generation consists of five epochs.

\subsection{Comparison with State-of-the-arts}

We compare our approach with other state-of-the-art methods NetVLAD \cite{netvlad}, CRN \cite{crn}, SARE \cite{sare}, and SFRS \cite{sfrs} on the Pitts250k-test and Tokyo 24/7 datasets (Tables \ref{pitts} and \ref{tokyo}, respectively). 

NetVLAD: embeds the traditional VLAD structure into the CNN network structure and can be easily applied to any CNN structure. 

CRN: (Contextual Reweighting Network) is a way of learning weights based on image context. By adding spatial attention, the network can focus on those regions that contribute positively to geo-localization.

SARE: (The Random Attractive and Repulsive Embedding) loss function is proposed on top of the VLAD aggregated feature embedding in SARE. 

SFRS: (The Self-supervising Fine-grained Region Similarities) improves NETVLAD by proposing self-supervised learning of image-to-region similarities. 

None of the above methods solves the bottleneck of self-adaptation and robustness in image localization training well.

Our proposed network achieves 85.7\% rank-1 recall on challenging Tokyo 24/7 dataset, outperforming the second-best performing method by 2.5\%. Our method also achieves state-of-the-art recall on the Pitts250 dataset, improving all reported recalls by around 1\%.

\begin{table}[htbp]
\caption{Comparison with state-of-the-arts on image-based localization benchmarks} \label{pitts}
\begin{center}
\begin{tabular}{|c|c|c|c|c|}
\hline
\multirow{2}{*}{\textbf{Method}} & \multirow{2}{*}{\textbf{Year}} & \multicolumn{3}{c|}{\textbf{Pitts250k}}   \\
\cline{3-5}
&                       & recall@1 & recall@5 & recall@10 \\
\hline 
NetVLAD                 & 2016                  & 86       & 93.2     & 95.1      \\
\hline
CRN                     & 2017                  & 85.5     & 93.5     & 95.5      \\
\hline
SARE                    & 2019                  & 89       & 95.5     & 96.8      \\
\hline
SFRS                    & 2020                  & 89.8     & 95.5     & 96.8      \\
\hline
\textbf{Ours}           & \textbf{2021}         & \textbf{90.7}     & \textbf{96.1}     & \textbf{97.3}      \\
\hline
\end{tabular}
\end{center}
\end{table}

\begin{table}[htbp]
\caption{Comparison with state-of-the-arts on image-based localization benchmarks} \label{tokyo}
\begin{center}
\begin{tabular}{{|c|c|c|c|c|}}
\hline
\multirow{2}{*}{\textbf{Method}} & \multirow{2}{*}{\textbf{Year}} &\multicolumn{3}{c|}{\textbf{Tokyo 24/7}}   \\
\cline{3-5}
&                       & recall@1 & recall@5 & recall@10 \\
\hline
NetVLAD                 & 2016                  & 73.3       & 82.9     & 86      \\
\hline
CRN                     & 2017                  & 75.2     & 83.8     & 87.3      \\
\hline
SARE                    & 2019                  & 79.7       & 86.7     & 90.5      \\
\hline
SFRS                    & 2020                  & 83.2     & 89.5     & 91.1      \\
\hline
\textbf{Ours}           & \textbf{2021}         & \textbf{85.7}     & \textbf{91.4}     & \textbf{93.7}      \\
\hline
\end{tabular}
\end{center}
\end{table}

We further visualize and inspect the top results obtained with our model and compare them to SFRS. Three examples are shown in Fig. \ref{Fig. 3}. In the first example, SFRS mistakenly focuses on the similarity of the wall area, while our method learns to ignore these misleading local areas which are not discriminative across a wide range of augmentation policies. In the second example, our method pays more attention to the positional relationship of vegetation and buildings, which provides valuable information for localization in urban-scale street scenes. Although both methods fail on the third example, the top-1 image we retrieved is more visually similar to the query image.

\begin{figure}[h]
\centering 
\includegraphics[width=0.5\textwidth]{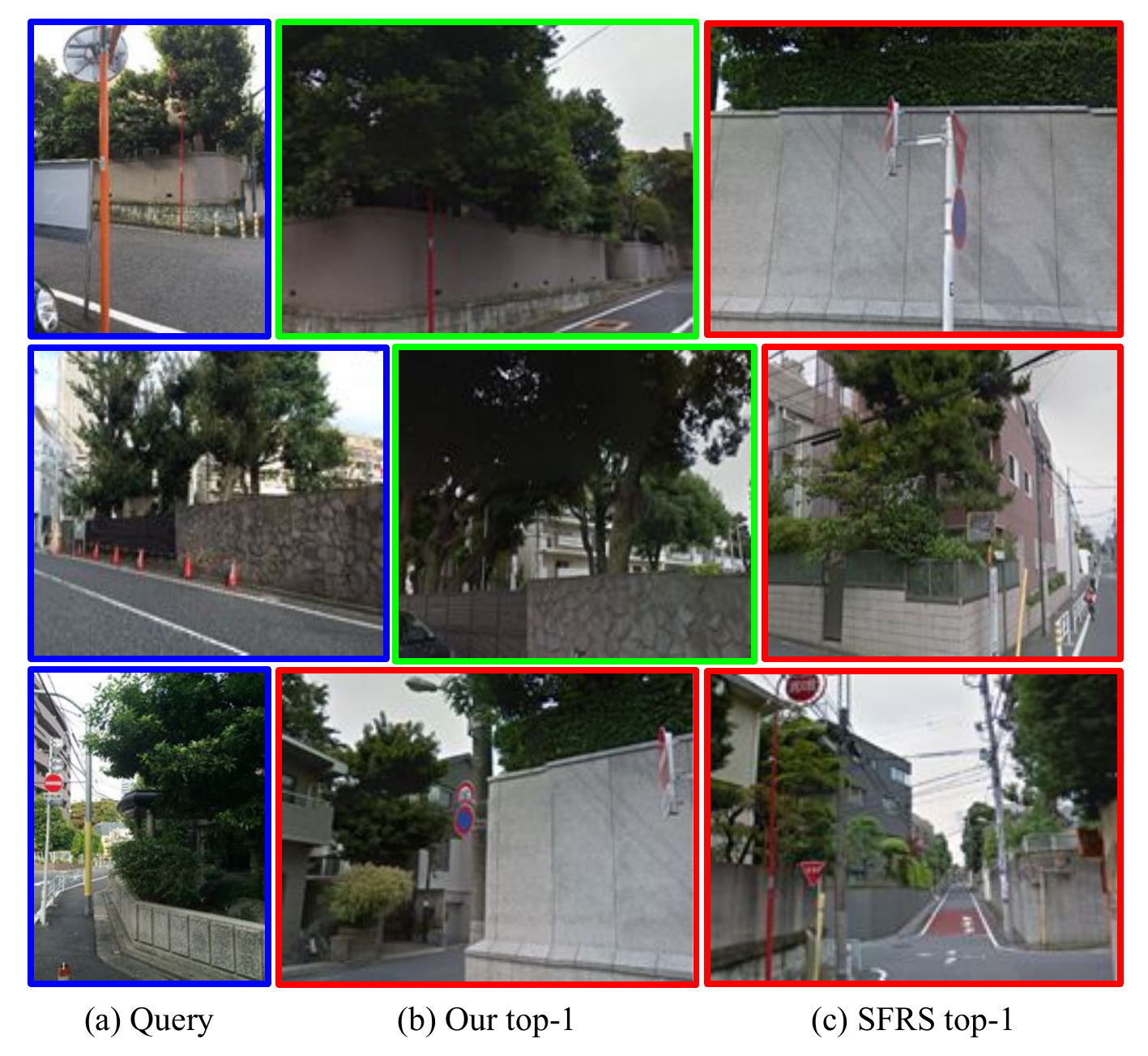} 
\caption{Example retrieval results on tokyo 24/7 dataset. From left to right: query image, the top retrieved image using our method, the top retrieved image using SFRS. Green and red borders indicate correct and incorrect retrieved results, respectively.} 
\label{Fig.3} 
\end{figure}

\subsection{Generalization on Image Retrieval Datasets}
 In order to assess the generalization power of our method we evaluate it on the rOxford and rParis benchmarks without any fine-tuning and focus on the more challenging Medium (M) and Hard (H) protocols. Table \ref{generalization} reports the performance of our method and other state-of-the-art IR methods. Our method improves the current state-of-the-art mAP by a noticeable margin on both datasets. Specifically, compared with baseline \cite{sfrs}, it achieves a +45\% improvement on rOxford(H) and a 9.1\% on rParis(H). Notably, although the proposed network is trained using a large-scale place recognition dataset from urban areas, it generalizes well to natural landmarks, leading to a boost in recognition performance.

\begin{table}[htbp]
\caption{Comparison with state-of-the-art methods on the rOxford and rParis benchmarks. All methods are applied without fine-tuning. M and H denote Medium and Hard retrieval protocols, respectively.} \label{generalization}
\begin{center}
\begin{tabular}{|c|c|c|c|c|c|}
\hline
\multirow{2}{*}{\textbf{Method}} & \multirow{2}{*}{\textbf{Year}} & \multicolumn{2}{c|}{\textbf{rOxford}} & \multicolumn{2}{c|}{\textbf{rParis}}     \\
\cline{3-6}
&                       & M & H & M & H\\
\hline 
DenseVLAD                 & 2016                  & 36.8 & 13 & 42.5 & 13.7\\
\hline
NetVLAD                   & 2017                  & 37.1 & 13.8 & 59.8 & 35.0      \\
\hline
SARE                    & 2019                  & -       & -     & -       & -       \\
\hline
SFRS                    & 2020                  & 43.2     & 17.2     & 57.3 & 30.6     \\
\hline
\textbf{Ours}           & \textbf{2021}         & \textbf{47.4}     & \textbf{25.0}     & \textbf{59.2} & \textbf{33.7}     \\
\hline
\end{tabular}
\end{center}
\end{table}

\subsection{Ablation Studies}
In order to evaluate the contribution and effectiveness of our adversarial augmentation method, we perform an ablation study on the rOxford and rParis datasets (Table \ref{ablation}). We follow a similar protocol to the ablation experiments in AutoAugment \cite{autoaug} and consider four augmentation policies, using the same IR network architecture (SFRS architecture, used in our method as well):

Baseline: SFRS\cite{sfrs}, which is trained with standard data augmentation.

Fixed: Augmenting batch with standard data augmentation fixed throughout the training.

Random: Randomizing the operations and magnitudes in the augmentation policy.

Ours: Our proposed method, which is trained with adversarial policies and image retrieval process.

We observe that the mAP of the Fixed method is 0.2-3.5\% better than the Baseline, suggesting that augmenting each batch by fixing the data augmentation during the training is beneficial. In addition, we find that the mAP of the Random method achieves a further 1.5\% improvement, highlighting the benefit of random policy sampling. Finally, our proposed approach can effectively learn by auto-augmenting samples with adversarial policies as extra augmentation, boosting the rank-1 recall to 25\% and 33.4\% on rOxford(H) and rParis(H), respectively.

\begin{table}[htbp]
\label{ablation}
\caption{Ablation study for augmentation policy, evaluated on the rOxford and rParis benchmarks.  M and H denote Medium and Hard retrieval protocols, respectively.} 
\begin{center}
\begin{tabular}{|c|c|c|c|c|c|}
\hline

\multirow{2}{*}{\textbf{Method}} & \multirow{2}{*}{\textbf{Aug.policy}} & 
\multicolumn{2}{c|}{\textbf{rOxford}} & \multicolumn{2}{c|}{\textbf{rParis}}    \\
\cline{3-6}
&                       & M & H & M & H\\
\hline 
Baseline                 & standard                   & 43.2     & 17.2     & 57.3 & 30.6\\
\hline 
Fixed                   & standard                  & 44.3 & 20.7 & 57.5 & 31.5      \\
\hline 
Random                    & random                  & 45.8      &21.5     &58.1       & 30.5       \\    
\hline
\textbf{Ours}           & \textbf{adversarial}         & \textbf{47.4}     & \textbf{25.0}     & \textbf{59.0} & \textbf{33.4}     \\
\hline 
\end{tabular}
\end{center}
\end{table}

\section{CONCLUSION}
In this work, we propose a novel adversarial framework for training IR networks. We formulate data augmentation policy search as a min-max game with an IR network, as a method to encourage robust and discriminative image representation learning. Our method overcomes the limitations of current IR methods for place recognition, and adaptively generates hard positives not seen in the data, based on the retrieval process. Results of extensive experiments on public datasets show that our method is effective and exhibits state-of-the-art performance for place recognition.
\section{Acknowledgement}
This work was supported by the Key-Area Research and 
Development Program of Guangdong Province 
(2020B0909050003), and Science and Technology 
Innovation Committee of Shenzhen 
(CJGJZD20200617102801005).


\bibliography{ref} 

\end{document}